\documentclass{article} 
\usepackage{iclr2022_conference,times}

\usepackage{amsmath,amsfonts,bm}

\def\eqref#1{equation~\ref{#1}}

\def\1{\bm{1}}

\DeclareMathAlphabet{\mathsfit}{\encodingdefault}{\sfdefault}{m}{sl}
\SetMathAlphabet{\mathsfit}{bold}{\encodingdefault}{\sfdefault}{bx}{n}

\usepackage{hyperref}
\usepackage{url}
\usepackage{booktabs}       
\usepackage{amsfonts}       
\usepackage{nicefrac}       
\usepackage{microtype}      
\usepackage{xcolor}         
\usepackage{amsfonts}       
\usepackage{float}
\usepackage{sidecap}
\usepackage{tikz}
\usepackage{amsmath}
\usepackage{wrapfig}

\title{Memory-Driven Text-to-Image Generation}

\author{%
  Bowen Li, Philip H. S. Torr, Thomas Lukasiewicz\\
  University of Oxford
  }

\iclrfinalcopy 
\begin{document}

\maketitle

\begin{abstract}

We introduce a memory-driven semi-parametric approach to text-to-image generation, which is based on both parametric and non-parametric techniques. The non-parametric component is a memory bank of image features constructed from a training set of images. The parametric component is a generative adversarial network. Given a new text description at inference time, the memory bank is used to selectively retrieve image features that are provided as basic information of target images, which enables the generator to produce realistic synthetic results. We also incorporate the content information into the discriminator, together with semantic features, allowing the discriminator to make a more reliable prediction. Experimental results demonstrate that the proposed memory-driven semi-parametric approach produces more realistic images than purely parametric approaches, in terms of both visual fidelity and text-image semantic consistency.

\end{abstract}

\vspace{-2ex}
\section{Introduction}
\vspace{-1ex}
How to effectively produce realistic images from given natural language descriptions with semantic alignment has drawn much attention, because of its tremendous potential applications in art, design, and video games, to name a few. Recently, with the vast development of generative adversarial networks~\citep{goodfellow2014generative, gauthier2015conditional, mirza2014conditional} in realistic image generation, text-to-image generation has made much progress, where the progress has been mainly driven by parametric models –- deep networks use their weights to represent all data concerning realistic appearance~\citep{zhang2017stackgan, zhang2018stackgan++, xu2018attngan, li2019controllable, qiao2019mirrorgan, zhu2019dm, hinz2019semantic, cheng2020rifegan, qiao2019learn}.

\begin{figure}[t]
\begin{minipage}{1\textwidth}
\begin{minipage}{0.18\textwidth}
\centering
\small{A zebra is standing on the grassy field.}
\end{minipage}
\;\;\begin{minipage}{0.195\textwidth}
\includegraphics[width=1\linewidth, height=1\linewidth]{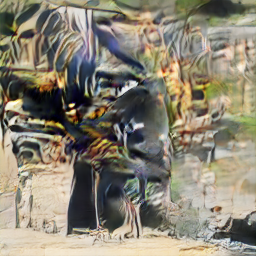}
\end{minipage}
\noindent\begin{minipage}{0.195\textwidth}
\includegraphics[width=1\linewidth, height=1\linewidth]{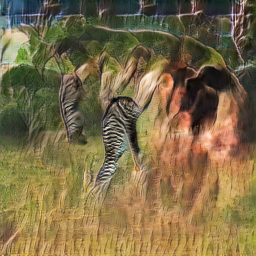}
\end{minipage}
\noindent\begin{minipage}{0.195\textwidth}
\includegraphics[width=1\linewidth, height=1\linewidth]{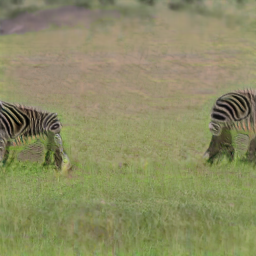}
\end{minipage}
\noindent\begin{minipage}{0.195\textwidth}
\includegraphics[width=1\linewidth, height=1\linewidth]{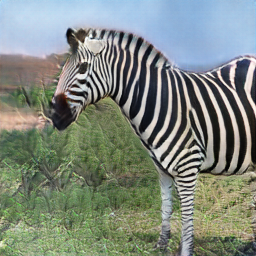}
\end{minipage}
\end{minipage}
\smallskip

\begin{minipage}{1\textwidth}
\begin{minipage}{0.18\textwidth}
\centering
\small{A white and blue bus is driving down a street.}
\end{minipage}
\;\;\begin{minipage}{0.195\textwidth}
\includegraphics[width=1\linewidth, height=1\linewidth]{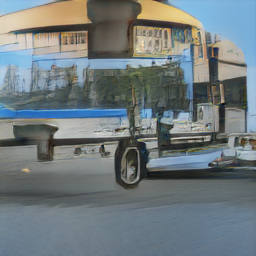}
\end{minipage}
\noindent\begin{minipage}{0.195\textwidth}
\includegraphics[width=1\linewidth, height=1\linewidth]{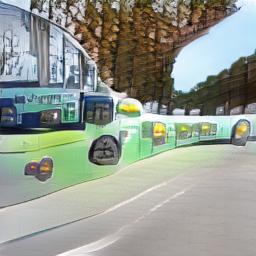}
\end{minipage}
\noindent\begin{minipage}{0.195\textwidth}
\includegraphics[width=1\linewidth, height=1\linewidth]{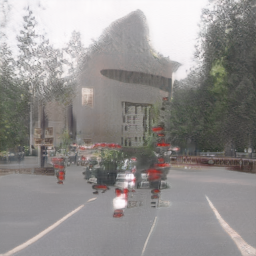}
\end{minipage}
\noindent\begin{minipage}{0.195\textwidth}
\includegraphics[width=1\linewidth, height=1\linewidth]{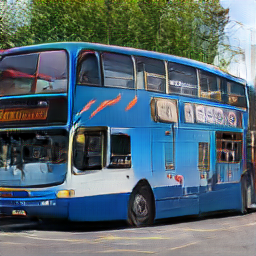}
\end{minipage}
\smallskip

\begin{minipage}{1\textwidth}
\begin{minipage}{0.195\textwidth}
\centering
\small{Given Text}
\end{minipage}
\begin{minipage}{0.195\textwidth}
\centering
\small{StackGAN++~\citep{zhang2018stackgan++}}
\end{minipage}
\noindent\begin{minipage}{0.195\textwidth}
\centering
\small{AttnGAN~\citep{xu2018attngan}}
\end{minipage}
\noindent\begin{minipage}{0.195\textwidth}
\centering
\small{DF-GAN~\citep{tao2020df}}
\end{minipage}
\noindent\begin{minipage}{0.195\textwidth}
\centering
\small{Ours}
\end{minipage}
\end{minipage}
\end{minipage}

\centering\vspace{-1ex}
\caption{Examples of text-to-image generation on COCO. Current approaches only generate low-quality images with unrealistic objects. In contrast, our method can produce realistic images, in terms of both visual appearances and geometric structure.}
\label{fig:intro}
\end{figure}

Although these approaches can produce realistic results on well-structured datasets, containing a specific class of objects at the image center with fine-grained descriptions,
such as birds~\citep{wah2011caltech} and flowers~\citep{nilsback2008automated}, there is still much room to improve. Besides, they usually fail on more complex datasets, which contain multiple objects with diverse backgrounds, e.g., COCO~\citep{lin2014microsoft}. 
This is likely because, for COCO, the generation process involves a large variety in objects (e.g., pose, shape, and location), backgrounds, and scenery settings. Thus, it is much easier for these approaches to only produce text-semantic-matched appearances instead of capturing difficult geometric structure. As shown in Fig.~\ref{fig:intro}, current approaches are only capable of producing required appearances semantically matching the given descriptions (e.g., white and black stripes for zebra), but objects are unrealistic with distorted shape.  
Furthermore, these approaches are in contrast to earlier works on image s, which were based on non-parametric techniques that could make use of large datasets of images at inference time~\citep{chen2009sketch2photo, hays2007scene, isola2013scene, zhu2015learning, lalonde2007photo}. 
Although parametric approaches can enable the benefits of end-to-end training of highly expressive models, they lose a strength of earlier non-parametric techniques, as they fail to make use of large datasets of images at inference time.

In this paper, we introduce a memory-driven semi-parametric approach to text-to-image generation, where the approach takes the advantage of both parametric and non-parametric techniques. The non-parametric component is a memory bank of disentangled image features constructed from a training set of real images. The parametric component is a generative adversarial network. Given a novel text description at inference time, the memory bank is used to selectively retrieve compatible image features that are provided as basic information, allowing the generator to directly draw clues of target images, and thus to produce realistic synthetic results.

Besides, to further improve the differentiation ability of the discriminator, we incorporate the content information into it. This is because, to make a prediction, the discriminator usually relies on semantic features, extracted from a given image using a series of convolution operators with local receptive fields. However, when the discriminator goes deeper, less content details are preserved, including the exact geometric structure information~\citep{gatys2016image, johnson2016perceptual}. We think that the loss of content details is likely one of the reasons why current approaches fail to produce realistic shapes for objects on difficult datasets, such as COCO. Thus, the adoption of content information allows the model to exploit the capability of content details and then improve the discriminator to make the final prediction more reliable. 

Finally, an extensive experimental analysis is performed, which demonstrates that our memory-driven semi-parametric method can generate more realistic images from natural language, compared with purely parametric models, in terms of both visual appearances and geometric structure.


\vspace{-2ex}
\section{Related Work}
\vspace{-1ex}

\textbf{Text-to-image generation} has made much progress because of the success of generative adversarial networks (GANs)~\citep{goodfellow2014generative} in realistic image generation. \cite{zhang2017stackgan} proposed a multi-stage architecture to generate realistic images progressively. Then, attention-based methods~\citep{xu2018attngan, li2019controllable} are proposed to further improve the results. \cite{zhu2019dm} introduced a dynamic memory module to refine image contents. \cite{qiao2019learn} proposed text-visual co-embeddings to replace input text with corresponding visual features. \cite{cheng2020rifegan} introduced a rich feature generating text-to-image synthesis.
Besides, extra information is adopted on the text-to-image generation process, such as scene graphs~\citep{johnson2018image, ashual2019specifying} and layout (e.g., bounding boxes or segmentation masks)~\citep{hong2018inferring, li2019object, hinz2019semantic}. 
However, none of the above approaches adopt non-parametric techniques to make use of large datasets of images at inference time, neither feed content information into the discriminator to enable a finer training feedback. Also, our method does not make use of any additional semantic information, e.g., scene graphs and layout.

\textbf{Text-guided image manipulation} is related to our work, where the task also takes natural language descriptions and real images as inputs, but it aims to modify the images using given texts to achieve semantic consistency~\citep{nam2018text, dong2017semantic, li2020manigan}. Differently from it, our work focuses mainly on generating novel images, instead of editing some attributes of the given images. Also, the real images in the text-guided image manipulation task behave as a condition, where the synthetic results should reconstruct all text-irrelevant attributes from the given real images. Differently, the real images in our work are mainly to provide the generator with additional cues of target images, in order to ease the whole generation process.

\textbf{Memory Bank.} \cite{qi2018semi} introduced a semi-parametric approach to realistic image generation from semantic layouts. \cite{li2019pastegan} used the stored image crops to determine the appearance of objects. \cite{tseng2020retrievegan} used a differentiable retrieval process to select mutually compatible image patches. \cite{li2021controllable} studied conditional image extrapolation to synthesize new images guided by the input structured text. Differently, instead of using a concise semantic representation (a scene graph as input), which is less user-friendly and has limited context of general descriptions, we use natural language descriptions as input. Also, \cite{liang2020cpgan} designed a memory structure to parse the textual content. Differently, our method simply uses a deep network to extract image features, instead of involving complex image preprocessing to build a memory bank.

\vspace{-2ex}
\section{Overview}
\vspace{-2ex}
\begin{wrapfigure}{r}{0.50\textwidth}
  \begin{center}
    \includegraphics[width=1\linewidth, height=0.37\linewidth]{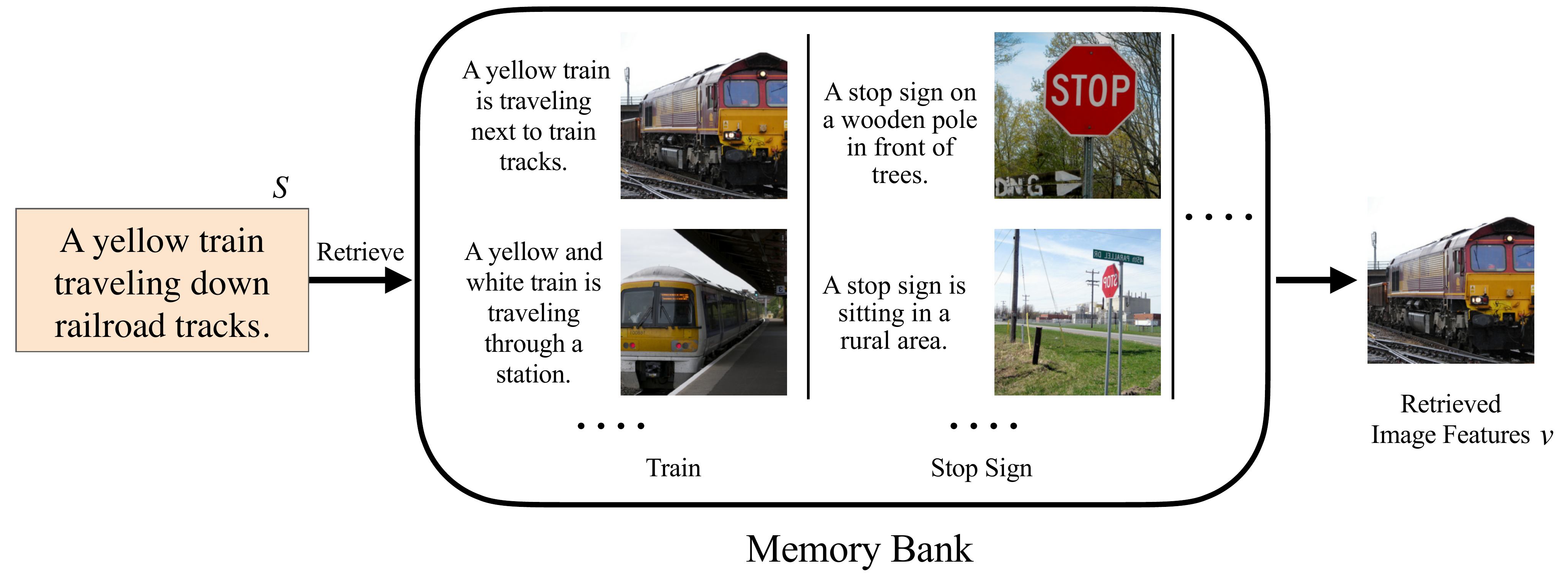}
  \end{center}\vspace{-3ex}
  \caption{The design of the memory bank to provide image features at inference time. Note that we use the corresponding real training image to represent image features for a better visualization.}
  \label{fig:archi_memory}
\end{wrapfigure}
Given a sentence $S$, we aim to generate a fake image $I'$ that is semantically aligned with the given $S$. The proposed model is trained on a set of paired text description and corresponding real image features $v$, denoted by $(S,v)$. This set is also used to generate a memory bank $M$ of disentangled image features $v$ for different categories, where image features are extracted from the training image by using a pretrained VGG-16 network~\citep{simonyan2014very} (see Fig.~\ref{fig:archi_memory}). Each element in $M$ is an image feature extracted from a training image, associated with corresponding semantically-matched text descriptions from the training datasets.

\begin{figure}[t]
\centering
\begin{minipage}{1\textwidth}
\includegraphics[width=1\linewidth, height=0.3218\linewidth]{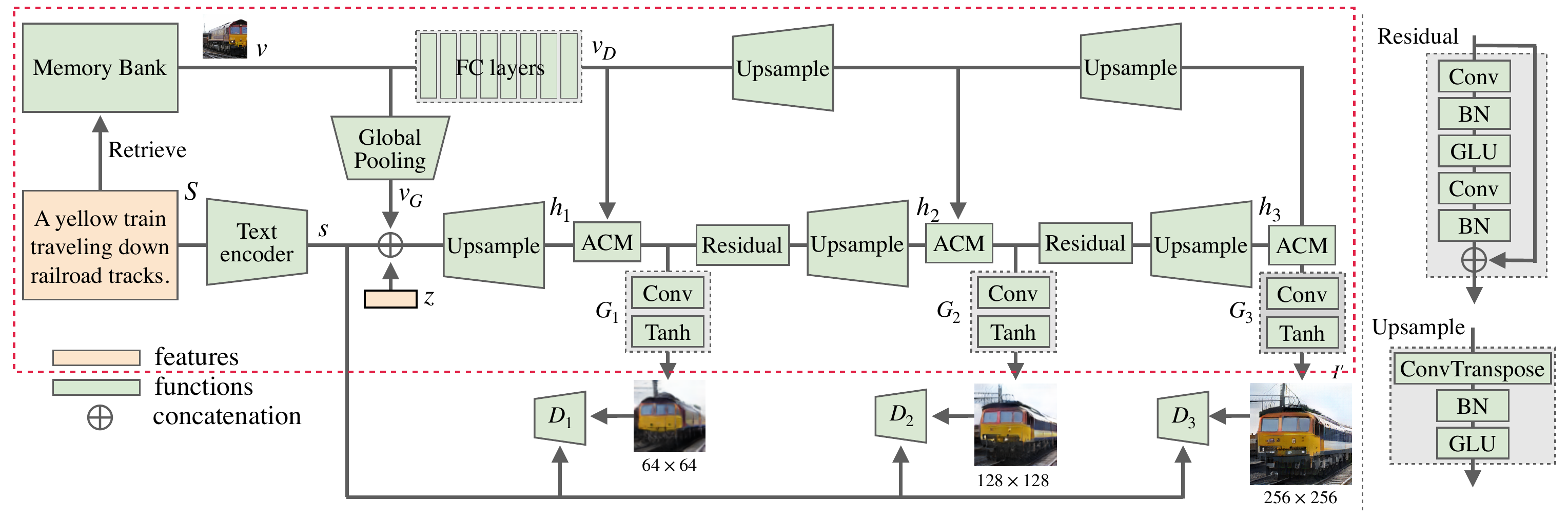}
\end{minipage}\vspace{-3ex}
\caption{The architecture of our proposed method. The red box indicates the inference pipeline that retrieves image features from a memory bank according to the a given description $S$, during in training, we directly feed image features from the text-paired training image. 
$z$ is a random vector drawn from the Gaussian distribution, ACM denotes the text-image affine combination module.}
\label{fig:archi}
\end{figure}

At inference time, we are given a novel text description $S$ that was not seen during training. Then, $S$ is used to retrieve semantically-aligned image features from the memory bank $M$, based on designed matching algorithms (more details are shown in Sec.~\ref{sec:retrieval}). Next, the retrieved image features~$v$, together with the given text description $S$, are fed into the generator to synthesize the output image (see Fig.~\ref{fig:archi}). The generator utilizes the information from the image features, fuses them with hidden features produced from the given text description $S$, and generate realistic images semantically-aligned with $S$. The architecture and training of the network are described in Sec.~\ref{sec:gan}.

To incorporate image features into the generation pipeline, we borrow from the text-guided image manipulation literature~\citep{li2020manigan}, and redesign the architecture to make full use of the given image features in text-to-image generation, shown in Fig.~\ref{fig:archi}. 

%
\vspace{-2ex}
\section{Memory Bank}
\vspace{-2ex}
\subsection{Representation}
\vspace{-1ex}
The memory bank $M$ is a set of image features $v_{i}$ extracted from training set images, and each image features $v_i$ is associated with matched text descriptions that are provided in the dataset, e.g., in COCO, each image has five matched text descriptions. These descriptions are used in the matching algorithms, allowing a given text to find the most compatible image features at inference time. 

\subsection{Retrieval}
\label{sec:retrieval}
\vspace{-1ex}
Given a new text description, in order to effectively retrieve the most compatible image features from the memory bank $M$, we have designed several matching algorithms and also explored the effectiveness of each algorithms. A detailed comparison between different algorithms is shown in the supplementary material.

\subsubsection{Sentence-Sentence Matching}
\label{sec:sent_sent}
\vspace{-1ex}
Here, we use image features' associated sentences $S'_i$ as keys, to find the most compatible image features $v_i$ for a given unseen sentence $S$ at inference time. First, we feed both $S$ and $S'_i$ into a pretrained text encoder~\citep{xu2018attngan} to produce sentence features $s \in \mathbb{R}^{D \times 1}$ and $s'_i \in \mathbb{R}^{D \times 1}$, respectively, where $D$ is the feature dimension. Then, for the given sentence $S$, we select the most compatible image features $v_i$ in $M$ based on a cosine similarity score:
\begin{equation}
    \alpha_{i}=\frac{(s)^{T}s'_{i}}{\left \| s \right \|\left \| s'_{i} \right \|}.
\end{equation}
Finally, we fetch the image features $v_i$ using the key $S'_{i}$ with the highest similarity score $\alpha_i$.

\subsubsection{Sentence-Image Matching}
\label{sec:sent_image}
\vspace{-1ex}
Instead of using associated sentences as keys, we can calculate the similarity between the sentence feature $s \in \mathbb{R}^{D \times 1}$ and image features $v_{i} \in \mathbb{R}^{D \times H \times W}$ stored in $M$, where $D$ is the number of channels, $H$ is the height, and $W$ is the width. To directly calculate the similarity, we first average the image features on the spatial direction to get a global image feature $v_{Gi} \in \mathbb{R}^{D \times 1}$. So, for a given unseen $S$, we select the most compatible image features $v_i$ in $M$ based on $\beta_i$:
\begin{equation}
    \beta_{i}=\frac{(s)^{T}v_{Gi}}{\left \| s \right \|\left \| v_{Gi} \right \|}.
\end{equation}
\subsubsection{Words-Words Matching}
\label{sec:word_word}
\vspace{-1ex}
Moreover, we can use a more fine-grained text representation (namely, word embeddings), as keys to find the most compatible image features $v_i$ stored in $M$ for a given unseen sentence $S$. 
At inference time, we first feed both $S$ and $S'_i$ into a pretrained text encoder~\citep{xu2018attngan} to generate word embeddings $w \in \mathbb{R}^{N \times D}$ and $w'_i \in \mathbb{R}^{N \times D}$, respectively, where $N$ is the number of words and $D$ is the feature dimension. Then, we reshape the size of both $w$ and $w'_i$ to $\mathbb{R}^{(D\ast N) \times 1}$.
So, to find the most compatible image features, the cosine similarity score can be defined as follows:
\begin{equation}
    \delta_{i}=\frac{(w)^{T}w'_{i}}{\left \| w \right \|\left \| w'_{i} \right \|}.
    \label{eq:word-word}
\end{equation}
However, different words in a sentence are not equally important. 
Thus, if we simply combine all words from a sentence together to calculate the similarity (like above), the similarity score may be less precise. 
To solve this issue, during training, we reweight each word in a sentence by its importance. We first use convolutional layers to remap word embeddings, and then calculate the importance $\lambda$ (and $\lambda'_i$) for each word in word embeddings $w \in \mathbb{R}^{N \times D}$ (and $w'_i \in \mathbb{R}^{N \times D}$), denoted by: $\lambda = \text{Softmax}(ww^{T})$ and $\lambda'_i = \text{Softmax}(w'_iw'^{T}_i)$, respectively. 

Each elements in $\lambda$ represents the correlation between different words in a sentence. Then, $\lambda w$ (and $\lambda'_i w'_{i}$) reweight word embeddings for each word based on its correlation with other words. So, using this reweighted word embeddings, we can achieve a more precise similarity calculation between two word embeddings.
At inference time, after we reshape the size of both $\lambda w$ and $\lambda'_i w'_i$ to $\mathbb{R}^{(D\ast N) \times 1}$, the new equation is defined as follows:
\begin{equation}
    \delta_{i}=\frac{(\lambda w)^{T}\lambda'_i w'_{i}}{\left \| \lambda w\right \|\left \| \lambda'_i w'_{i}\right \|}.
    \label{eq:new_word-word}
\end{equation}

\subsubsection{Words-Image Matching}
\label{sec:word_image}
\vspace{-1ex}
Furthermore, we use the word embeddings $w \in \mathbb{R}^{N \times D}$ and image features $v_i \in \mathbb{R}^{D \times H \times W}$ to directly calculate the similarity score between them. To achieve this, we first reshape the size of the image features to $v_i \in \mathbb{R}^{D \times (H \ast W)}$. Then, a correlation matrix $c_{i} \in \mathbb{R}^{N \times (H \ast W)}$ can be obtained via: $c_{i} = \text{Softmax}(wv_{i})$, where each element in $c_{i}$ represents the correlation between each word and each image spatial location. Then, a reweighted word embedding $\tilde{w_{i}} \in \mathbb{R}^{N \times D}$ containing image information can be achieved by $\tilde{w_{i}} = c_{i}v_{i}^{T}$. 
So, to find the most compatible image features, we first reshape the size of both $w$ and $\tilde{w_{i}}$ to $\mathbb{R}^{(D \ast N) \times 1}$, and the similarity score is defined as follows:
\begin{equation}
    \gamma_{i}=\frac{(w)^{T}\tilde{w_{i}}}{\left \| w \right \|\left \| \tilde{w_{i}} \right \|}.
    \label{eq:word-image}
\end{equation}
Similarly, we can also reweight word embeddings $w$ and image features $v_i$ based on their importance (see Sec.\ref{sec:word_word}) to achieve a more precise calculation.

\vspace{-1ex}
\section{Generative Adversarial Networks}
\label{sec:gan}
\vspace{-2ex}
To generate high-quality synthetic images from natural language descriptions, we propose to incorporate image features $v$, along with the given sentence $S$, into the generator. 
To incorporate image features into the generation pipeline, we borrow from the text-guided image manipulation literature~\citep{li2020manigan}, and redesign the architecture to make full use of the given image features in text-to-image generation, shown in Fig.~\ref{fig:archi}. 
\vspace{-1ex}
\subsection{Generator with Image Features}
\vspace{-1ex}

To avoid the identity mapping and also to make full use of image features $v$ in the generator, we first average $v$ on each channel to filter potential content details (e.g., overall spatial structure) contained in~$v$, getting a global image feature $v_{G}$, where $v_G$ only keeps basic information of the corresponding real image $I$, serving as basic image priors. By doing this, the model can effectively avoid copying and pasting from $I$, and greatly ensure the diversity of output results, especially on the first stage. This is because the following stages focus more on refining basic images produced by the first stage, according to adding more details and improving their resolution, shown in Fig.~\ref{fig:archi}.  

However, only feeding the global image feature $v_G$ at the beginning of the network, the model may fail to fully utilize the cues contained in the image features $v$. Thus, we further incorporate the image features $v$ at each stage of the network. The reason to feed image features $v$ rather than the global feature $v_G$ at the following stages is that $v$ contains more information about the desired output image, such as image contents and geometric structure of objects, where these details can work as candidate information for the main generation pipeline to select. To enable this regional selection effect, we adopt the text-image affine combine module (ACM)~\citep{li2020manigan}, which is able to selectively fuse text-required image information within $v$ into the hidden features $h$, where $h$ is generated from the given text description $S$. However, simply fusing image features $v$ into the generation pipeline may introduce constraints on producing diverse and novel synthetic results, because different image information (e.g., objects and visual attributes) in $v$ may be entangled, which means, for example, if the model only wants to generate one object, the corresponding entangled parts (e.g, objects and attributes) may be produced as well. This may cause an additional generation of text-irrelevant objects and attributes. 
Thus, to avoid these drawbacks, inspired by the study~\citep{karras2019style}, we use several fully connected layers to disentangle the image features $v$, getting disentangled image features $v_{D}$, which allows the model to disconnect relations between different objects and also attributes. 
By doing this, the model is able to prevent the constraints introduced by the image features $v$, and then selectively choose text-required image information within $v_D$, where this information is effectively disentangled without a strong connection. 

\textbf{Why does the generator with image features work better?}
Ideally, the generator produces a sample, e.g., an image, from a latent code, and the distribution of these samples should be indistinguishable from the training distribution, where the training distribution is actually drawn from the real samples in the training dataset. Based on this, incorporating image features from real images in training dataest into the generator allows the generator to directly draw cues of the desired distribution that it eventually needs to generate. 
Besides, the global feature $v_G$ and disentangled image features $v_D$ can provide basic information of target results in advance, and also work as candidate information, allowing the model to selectively choose text-required information without generating it by the model itself, and thus easing the whole generation process. To some extent, the global feature $v_G$ can be seen as the meta-data of target images, which may contain information about what kinds of objects to generate, e.g., zebra or bus, and $v_D$ is able to provides basic information of objects, e.g., the spatial structure like four legs and one head for the zebra and the rectangle shape for the bus. 

\vspace{-1ex}
\subsection{Discriminator with Content Information}
\vspace{-1ex}
\begin{wrapfigure}{r}{0.52\textwidth}
  \begin{center}
    \includegraphics[width=1\linewidth, height=0.366\linewidth]{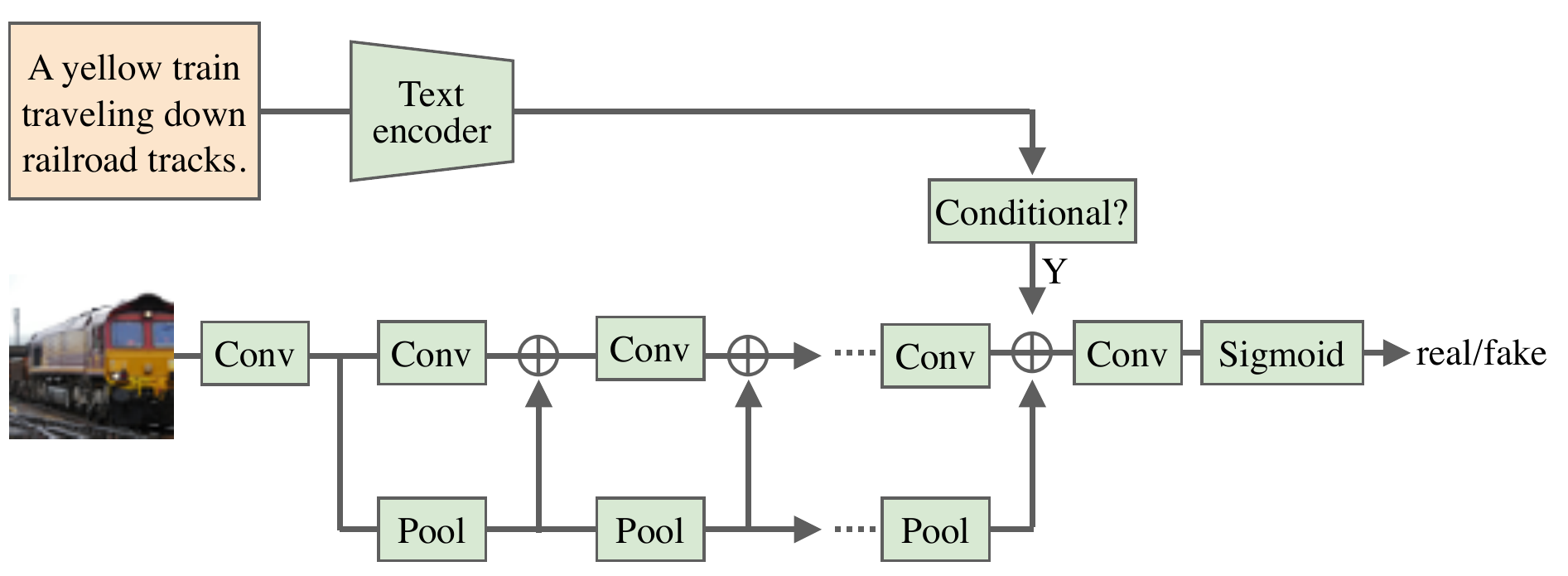}
  \end{center}\vspace{-3ex}
  \caption{The architecture of the proposed discriminator with the incorporation of content information.}
  \label{fig:archi_dis}
\end{wrapfigure}
To further improve the discriminator to make a more reliable prediction, with respect to both visual appearances and geometric structure, we propose to incorporate the content information into it. This is mainly because, in a deep convolution neural network, when the network goes deeper, the less content details are preserved, including the exact shape of objects~\citep{gatys2016image, johnson2016perceptual}. We think the loss of content details may prevent the discriminator to provide fine-grained shape-quality-feedback to the generator, which may cause the difficulty for the generator to produce realistic geometric structure. 
Also, \cite{zhou2014object} showed that the empirical receptive field of a deep convolution neural network is much smaller than the theoretical one especially on deep layers. This means, using convolution operators with a local receptive field only, the network may fail to capture the spatial structure of objects when the size of objects exceeds the receptive field. 

To incorporate the content details, we propose to generate a series of image content features, $\{a_{128}, a_{64}, a_{32},\ldots, a_{4}\}$, by aggregating different image regions via applying pooling operators on the given real or fake features. 
The size of these content features is from $a_{128} \in \mathbb{R}^{C \times 128 \times 128}$ to $a_{4} \in \mathbb{R}^{C \times 4 \times 4}$, where $C$ represents the number of channels, and the width and the height of the next image content features are $1/2$ the previous one. 
Thus, the given image is pooled into representations for different regions, from fine- ($a_{128}$) to coarse-scale ($a_{4}$), which is able to preserve content information of different subregions, such as the spatial structure of objects. Then, these features are concatenated with the corresponding hidden features on the channel-wise direction, incorporating the content information into the discriminator.

The number of different-scale content features can be modified, which is dependent on the size of given images. These features aggregate different image subregions by repetitively adopting fixed-size pooling kernels with a small stride. Thus, these content features maintain a reasonable small gap for image information. For the type of pooling operation between max and average, we perform comparison studies to show the difference in Sec.~\ref{sec:ablation}.

\textbf{Why does the discriminator with content information work better?}
Basically, the discriminator in a generative adversarial network is simply a classifier~\citep{goodfellow2014generative}. It tries to distinguish real data from the data created by the generator (note that in our method, we implement the Minmax loss in the loss function, instead of the Wasserstein loss~\citep{arjovsky2017wasserstein}).
Also, the implementation of content information has shown its great effectiveness on classification~\citep{lazebnik2006beyond, he2015spatial} and semantic segmentation~\citep{liu2015parsenet, zhao2017pyramid}. Based on this, incorporating the content information into the discriminator is helpful, allowing the discriminator to make a more reliable prediction on complex datasets, especially for the datasets with complex image scenery settings, such as COCO. 

\vspace{-2ex}
\subsection{Training}
\label{sec:training}
\vspace{-1ex}

To train the network, we follow~\citep{li2020manigan} and adopt adversarial training. There are three stages in the model, and each stage has a generator network and a discriminator network. The generator and discriminator are trained alternatively by minimizing the generator loss $\mathcal{L}_{G}$ and discriminator loss $\mathcal{L}_{D}$. Please see the supplementary material for more details about training objectives. We only highlight some training differences compared with~\cite{li2020manigan}.

\textbf{Generator objective.}
The objective functions to train the generator are similar as in \citep{li2020manigan}, but, differently, the inputs for the generator are a pair of (S, v) and a noise $z$, denoted by $G_{i}(z,S,v)$, where $i$ indicates the stage number.

\textbf{Discriminator objective.}
To improve the convergence of our GAN-based generation model, the $R_{1}$ regularization~\citep{mescheder2018training} is adopted in the discriminator:
\begin{equation}
R_1({\psi}):=\frac{\gamma }{2}E_{p_{D}(x)}\left [ \left \| \bigtriangledown D_{\psi }(x) \right \|^{2} \right ]
\textrm{,}
\label{eq:regularizaiton}
\end{equation}
where $\psi$ represents parameter values of the discriminator.
\vspace{-1ex}
\section{Experiments}
\label{sec:experiments}
\vspace{-4ex}
\begin{table}[h!]
  \centering
  \caption{Quantitative comparison on CUB bird: Fréchet inception distance (FID) and R-precision (R-psr) of StackGAN++~\citep{zhang2018stackgan++}, AttnGAN~\citep{xu2018attngan}, ControlGAN~\citep{li2019controllable}, DM-GAN~\citep{zhu2019dm}, DF-GAN\citep{tao2020df}, and our method. For FID, lower is better, while for R-precision, alignment, and realism, higher is better.}
  \label{table:quan_bird}
  \smallskip
  \scalebox{1}{
  \begin{tabular}{ccccccc}
    \toprule
    \multicolumn{1}{c}{Matrix} &
    \multicolumn{1}{c}{StackGAN++} & 
    \multicolumn{1}{c}{AttnGAN} & 
    \multicolumn{1}{c}{ControlGAN} &
    \multicolumn{1}{c}{DM-GAN} &
    \multicolumn{1}{c}{DF-GAN} &
    \multicolumn{1}{c}{Ours} \\             
    \midrule
    FID & 15.30 & 23.98 & 13.92 & 16.09 & 14.81 & 10.49 \\
    \midrule
    R-psr & 46.67 & 67.82 & 69.33 & 72.31 & - & 73.87 \\
    \midrule
    Alignment (\%) & - & - & - & - & 43 & 57 \\
    \midrule
    Realism (\%) & - & - & - & - & 31 & 69 \\
    \bottomrule
  \end{tabular}
  }
  \vspace*{-2ex}
\end{table}
\vspace{-3ex}
\begin{table}[h!]
  \centering
  \caption{Quantitative comparison on COCO. Note that we also compare our method with OP-GAN~\citep{hinz2019semantic}, where OP-GAN adopts \textbf{bounding box} in their method.}
  \label{table:quan_coco}
  \smallskip
  \scalebox{0.9}{
  \begin{tabular}{ccccccc|c}
    \toprule
    \multicolumn{1}{c}{Matrix} &
    \multicolumn{1}{c}{StackGAN++} & 
    \multicolumn{1}{c}{AttnGAN} & 
    \multicolumn{1}{c}{ControlGAN} &
    \multicolumn{1}{c}{DM-GAN} &
    \multicolumn{1}{c}{DF-GAN} &
    \multicolumn{1}{c}{Ours} &
    \multicolumn{1}{c}{OP-GAN} \\
    \midrule
    FID & 81.59 & 32.32 & 33.58 & 32.64 & 21.42 & 19.47 & 24.70  \\
    \midrule
    R-prs & 71.88 & 85.47 & 82.43 & 88.56 & - & 90.32 & 89.01 \\
    \midrule
    Alignment (\%) & - & - & - & - & 29 & 71 & - \\
    \midrule
    Realism (\%) & - & - & - & - & 22 & 78 & -\\
    \bottomrule
  \end{tabular}
  }
  \vspace*{-3ex}
\end{table}

\begin{figure}[h!]
\begin{minipage}{1\textwidth}
\centering
\includegraphics[width=1\linewidth, height=0.2622\linewidth]{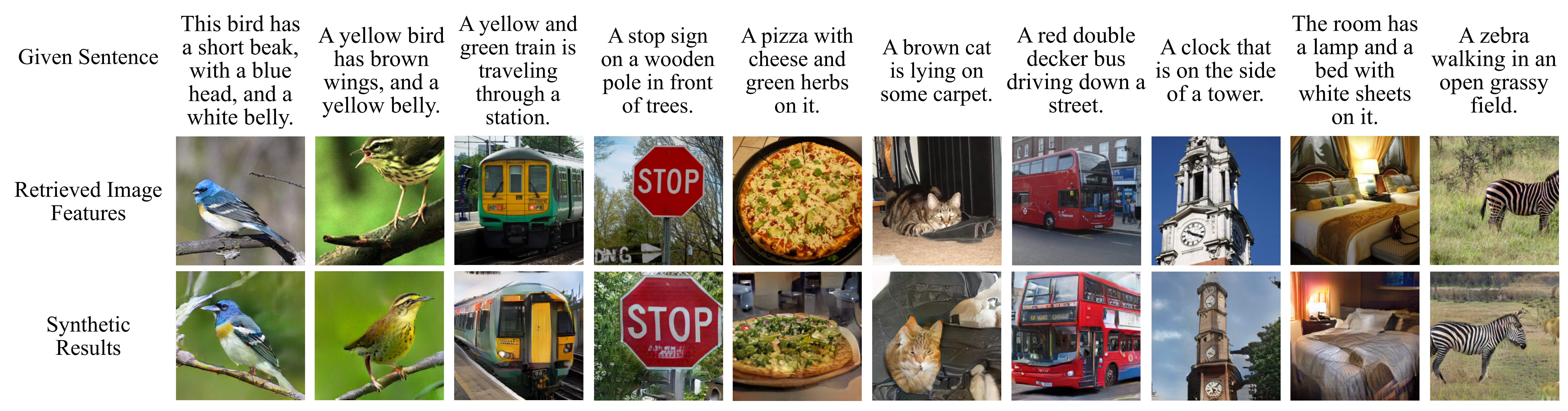}
\end{minipage}

\centering
\caption{Qualitative results on CUB and COCO: top row is the given unseen sentences; middle row: the image features extracted from the memory bank $M$ (we use corresponding images to represent the image features for a better visualization); bottom row: the synthetic results.}
\label{fig:qual_res}
\begin{minipage}{1\textwidth}
\centering
\includegraphics[width=1\linewidth, height=0.438\linewidth]{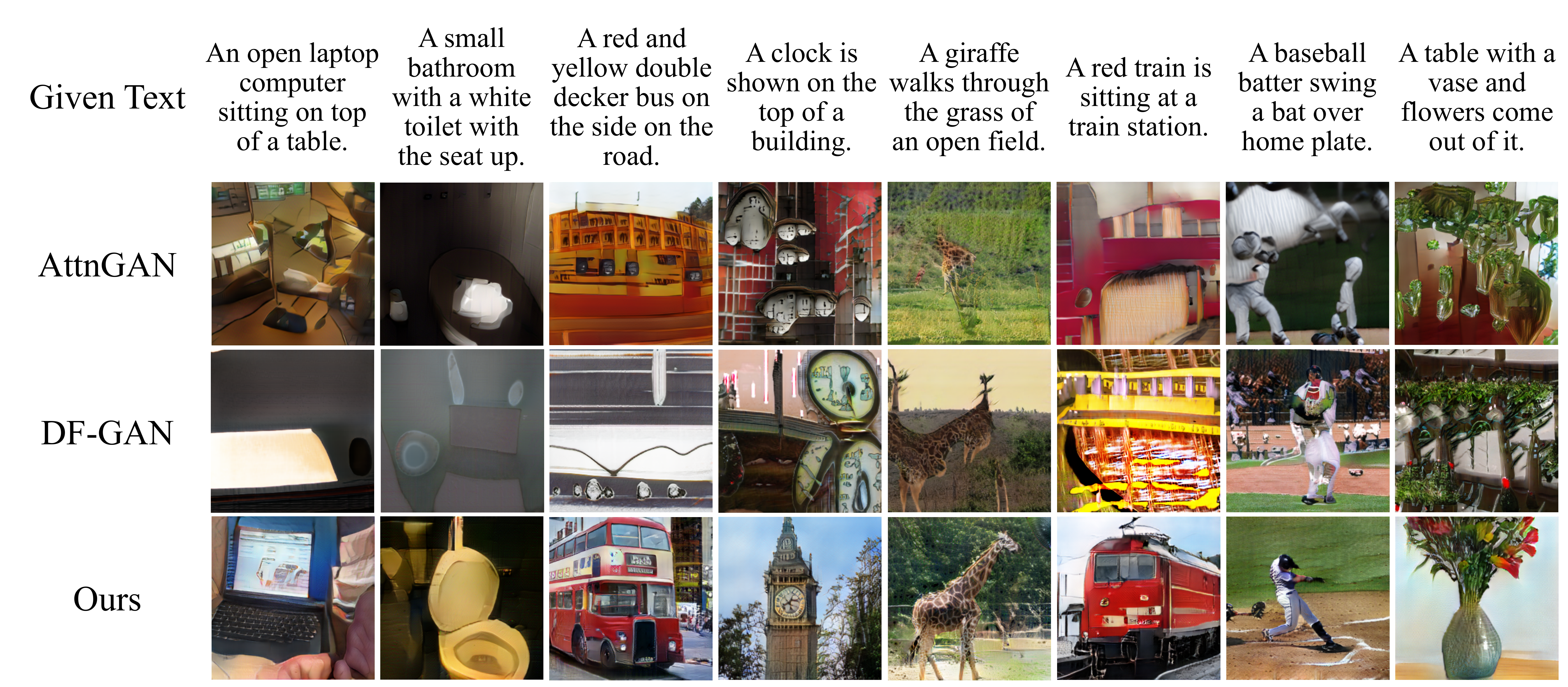}
\end{minipage}

\centering
\caption{Qualitative comparison between AttnGAN~\citep{xu2018attngan}, DF-GAN~\citep{tao2020df}, and our method on COCO.}
\label{fig:qual_cmp}
\vspace{-1ex}
\end{figure}

To verify the effectiveness of our proposed method in realistic image generation from text descriptions, we conduct extensive experiments on the CUB bird~\citep{wah2011caltech} dataset and more complex COCO~\citep{lin2014microsoft} dataset, where COCO contains multiple objects with diverse backgrounds.

\textbf{Evaluation metrics.} We adopt the Fréchet inception distance (FID)~\citep{heusel2017gans} as the primary metric to quantitatively evaluate the image quality and diversity. 
In our experiments, we use $30\text{K}$ synthetic images~vs.~$30\text{K}$ real test images to calculate the FID value. 
However, as FID cannot reflect the relevance between an image and a text description, we use the R-precision~\citep{xu2018attngan} to measure the correlation between a generated image and its corresponding text. 

\textbf{Human evaluation.}
To better verify the performance of our proposed method, we conducted a user study between current state-of-the-art method DF-GAN~\citep{tao2020df} and ours on CUB and COCO. We randomly selected 100 text descriptions from the test dataset. Then, we asked 5 workers to compare the results after looking at the output images and given text descriptions based on two criteria: (1) alignment: whether the synthetic image is semantically aligned with the given description, and (2) realism: whether the synthetic image looks realistic, shown in Tables~\ref{table:quan_bird} and~\ref{table:quan_coco}. Please see supplementary material for more details about the human evaluation.

\textbf{Implementation.} There are three stages in the model, and each stage has a generator network and a discriminator network. The number of stages can be modified, which depends on the resolution of the output image.
We utilize a deep neural network layer relu5\_3 of a pre-trained VGG-16 to extract image features $v$, which is able to filter content details in $I$ and keep more semantic information. 
In the discriminator, the number of different-scale image content features can be modified, which is related to the size of the given image. A same-size pooling kernel with a small stride ($\text{stride}=2$) is repeatedly implemented on the image features, to maximize the preservation of the content information. For the type of pooling operation, average pooling is adopted. For the matching algorithms, word image matching with reweighting based on importance is adopted. 
The resolution of synthetic results is $256\times 256$. Our method and its variants are trained on a single Quadro RTX 6000 GPU, using the Adam optimizer~\citep{kingma2014adam} with the learning rate $0.0002$. The hyperparameter $\lambda$ is set to~$5$. We preprocess datasets according to the method used in \citep{xu2018attngan}. \textbf{No} attention module is implemented in the whole architecture.

\subsection{Comparison with Other Approaches}
\vspace{-1ex}
\textbf{Quantitative comparison.}
Quantitative results are shown in Tables~\ref{table:quan_bird} and~\ref{table:quan_coco}. 
As we can see, compared to other approaches, our method achieves better FID and R-precision scores on both datasets, and even has a better performance than OP-GAN, where OP-GAN adopts bounding boxes. 
This indicates that (1) our method can produce more realistic images from given text descriptions, in terms of image quality and diversity, and (2) synthetic results produced by our method are more semantically aligned with the given text descriptions. Besides, in human evaluation, our method achieves better alignment and realism scores, compared with DF-GAN, which indicates that our results are most preferred by workers, which further verifies the better performance of our method, with respect to semantic alignment and image realism.

\begin{wrapfigure}{r}{0.50\textwidth}
\centering
\begin{minipage}{0.5\textwidth}
\centering
\includegraphics[width=1\linewidth, height=0.375\linewidth]{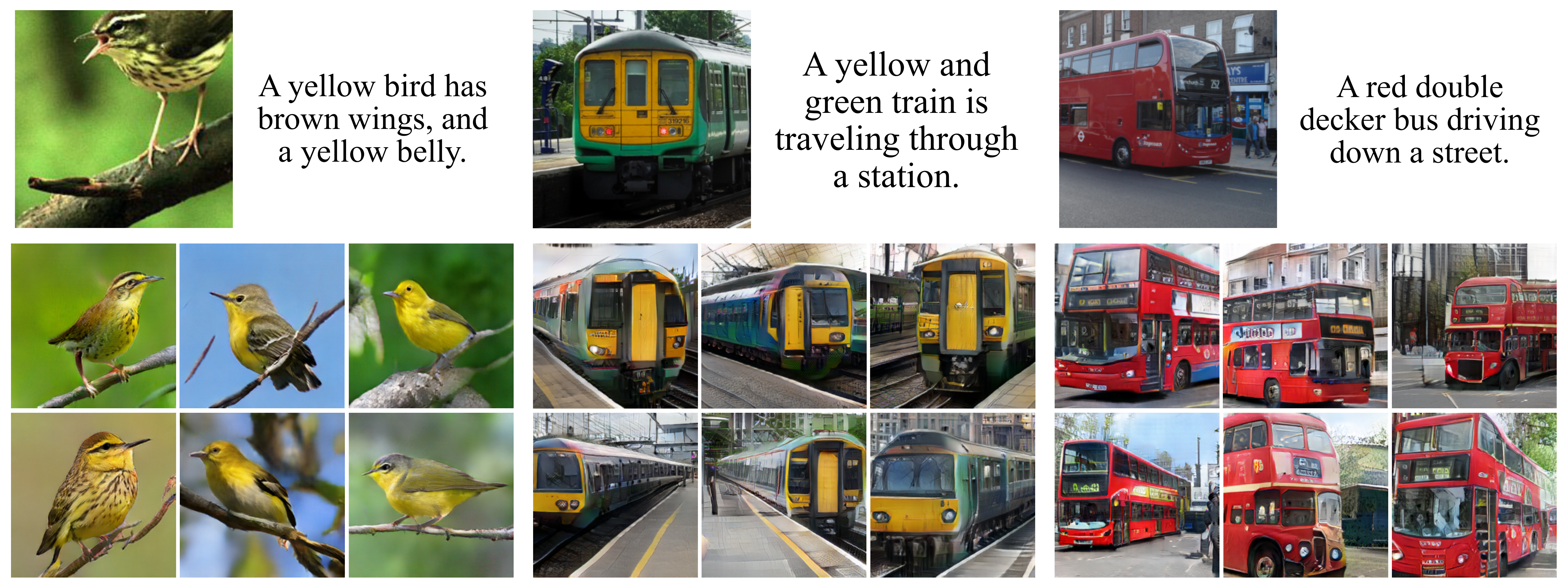}
\end{minipage}

\centering
\caption{Diversity. Top row shows the fixed sentence and image features, where we use the corresponding images to represent image features for a better visualization. The bottom presents diverse synthetic images produced by only changing the input noise $z$.}
\label{fig:diversity}
\end{wrapfigure}
\textbf{Qualitative comparison.}
In Fig.~\ref{fig:qual_res}, we present synthetic examples produced by our method at $256 \times 256$, along with the corresponding retrieved images that provide image features. As we can see, our method is able to produce high-quality results on CUB and COCO, with respect to realistic appearances and geometric structure, and also semantically matching the given text descriptions. Besides, the synthetic results are different from the retrieved image features, which indicates there is no significant copy-and-paste problem in our method.

\textbf{Diversity evaluation.} To further evaluate the diversity of our method, we fix the given text description and the corresponding retrieved image features, and only change the given noise $z$ to generate output images, shown in Fig.~\ref{fig:diversity}. When we fix the sentence and image features and only change the noise, our method can generate obviously different images, but they still semantically match the given sentence and also make use information from the image features. More evaluations are shown in the supplementary material.

\vspace{-1ex}
\subsection{Component Analysis}
\label{sec:ablation}
\vspace{-1ex}
\begin{wraptable}{r}{0.49\textwidth}
\vspace{-4.7ex}
  \centering
  \caption{Ablation studies: ``Ours w/o Feature'' denotes without feeding image features into the generator, ``Ours w/o Disen.'' denotes without using the fully connected layers to disentangle image features $v$, ``Ours w/o Disen.*'' is for mismatched pairs, ``Ours w/o Content'' denotes without incorporating the content information into the discriminator, ``Ours w/o Reg.'' denotes without using the regularization in the discriminator, ``Ours w/ Max'' denotes using the maximum pooling to extract content information, and ``Ours w/ Aver'' denotes using the average pooling.}
  \label{table:abla}
  \smallskip
  \scalebox{1}{
  \begin{tabular}{l||cc}
    \toprule
    Method & FID  & R-psr \\
    \midrule
    Ours w/o Feature & 22.20 & 84.63  \\
    Ours w/o Disen. & 18.82 & 92.17  \\
    Ours w/o Disen.* & 18.80 & 67.05  \\
    Ours w/o Content & 20.96 & 88.95 \\
    Ours w/o Reg. & 27.12 & 82.97  \\
    \midrule
    Ours w/ Max & 26.12 & 83.11  \\
    Ours w/ Aver (baseline) & 19.47 & 90.32  \\
    \bottomrule
  \end{tabular}
  }
\end{wraptable}

\textbf{Effectiveness of the image features.} 
To better understand the effectiveness of image features in the generator, we conduct an ablation study shown in Table~\ref{table:abla}. Without image features, the model ``Ours w/o Feature'' achieves worse quantitative results on both FID and R-precision compared with the baseline, which verifies the effectiveness of image features on high-quality image generation. Interestingly, without image features, even our method becomes a pure text-to-image generation method, similar to other baselines,  
but the FID of ``Ours w/o Feature'' is still competitive with other baselines. This indicate that even without the image features fed into our method, our method can still generate better synthetic results, with respect to image quality and diversity. We think this is mainly because with the help of content information, our better discriminator is able to make a more reliable prediction on complex datasets, which in turn encourages the generator to produce better synthetic images. 

\textbf{Effectiveness of the disentanglement.}
Here, we show the effectiveness of the fully connected layers applied on the image features $v$. Interestingly, from Table~\ref{table:abla}, the ``model w/o Disen.'' achieves better FID and R-precision compared with the baseline. This is likely because the model may suffer from an identity mapping problem. To verify this identity mapping problem, we conduct another experiment, where we feed mismatched sentence and image pairs into the network without using search algorithms, denoted ``model w/o Disen.*''. As we can see, on mismatched pairs, although FID is still low, the R-precision degrades significantly. 

\textbf{Effectiveness of the content information.}
To verify the effectiveness of the content information adopted in the discriminator, we conduct an ablation study, shown in Table~\ref{table:abla}. As we can see, FID and R-precision degrade when the discriminator without adopting the content information. 
This may indicate that the content information can effectively strengthens the differentiation abilities of the discriminator. Then, the improved discriminator is able to provide the generator with fine-grained training feedback, regarding to geometric structure, thus facilitating training a better generator to produce higher-quality synthetic results. 

\textbf{Comparison between different pooling types.}
Here, we conduct a comparison study on different pooling types (i.e., max and average) in Table~\ref{table:abla}. 
As we can see, the model with the average pooling works better than max pooling. We think that this is likely because max pooling fails to capture the contextual information between neighboring pixels, because it only picks the maximum value among a region of pixels, while average pooling calculates the average value between them. 

\textbf{Effectiveness of the regularization.}
We evaluate the effectiveness of the adopted regularization in the discriminator. From Table~\ref{table:abla}, the model without the regularization has worse quantitative results, compared with the full model. 
We think that this is because the regularization effectively improves GAN convergence by preventing the generator from training on junk feedback, once the discriminator cannot easily tell the difference between real and fake.

\vspace{-1ex}
\section{Conclusion}
\vspace{-1ex}
We have introduced a memory-driven semi-parametric approach to text-to-image generation, which utilizes large datasets of images at inference time. Also, an alternative architecture is proposed for both the generator and the discriminator. Extensive experimental results on two datasets demonstrate the effectiveness of feeding retrieved image features into the generator and incorporating content information into the discriminator.

\newpage
\bibliography{egbib}
\bibliographystyle{iclr2022_conference}

\end{document}